# TɪɢQA: An Expert-Annotated Question-Answering Dataset in Tigrinya

**Hailay Kidu Teklehaymanot**[*], **Dren Fazlija**[*], **Niloy Ganguly**[†]
**Gourab K. Patro**[*,†], **Wolfgang Nejdl**[*]

[*]L3S Research Center, Leibniz University Hannover, Germany
[†]Department of Computer Science and Engineering, IIT Kharagpur, India
{teklehaymanot, dren.fazlija, patro, nejdl}@l3s.de, niloy@cse.iitkgp.ac.in

**Abstract**

The absence of explicitly tailored, accessible annotated datasets for educational purposes presents a notable obstacle for NLP tasks in languages with limited resources. This study initially explores the feasibility of using machine translation (MT) to convert an existing dataset into a Tigrinya dataset in SQuAD format. As a result, we present TɪɢQA, an expert-annotated dataset containing 2,685 question-answer pairs covering 122 diverse topics such as climate, water, and traffic. These pairs are from 537 context paragraphs in publicly accessible Tigrinya and Biology books. Through comprehensive analyses, we demonstrate that the TɪɢQA dataset requires skills beyond simple word matching, requiring both single-sentence and multiple-sentence inference abilities. We conduct experiments using state-of-the-art MRC methods, marking the first exploration of such models on TɪɢQA. Additionally, we estimate human performance on the dataset and juxtapose it with the results obtained from pre-trained models. The notable disparities between human performance and the best model performance underscore the potential for future enhancements to TɪɢQA through continued research. Our dataset is freely accessible via the provided link to encourage the research community to address the challenges in the Tigrinya MRC. .
**Keywords:** Tigrinya QA dataset, Low resource QA dataset, domain specific QA

## 1. Introduction

The vast majority of human knowledge is documented through written text. Achieving a level of machine reading comprehension that closely resembles human understanding would open up a wide array of artificial intelligence applications. Like assessing human students' reading comprehension through questions based on text passages, machine reading comprehension (MRC) involves evaluating machines' understanding of written language by posing questions. These tests offer objective grading and can gauge various abilities, ranging from fundamental comprehension to causal reasoning and inference. The research community has adopted a methodology similar to machine reading comprehension (MRC) to teach machine literacy(Trischler et al., 2016a).

This task has become popular with the emergence of a large-scale and high-quality QA dataset namely SQuAD (Rajpurkar et al., 2016)leading to the release of other datasets such as SQuAD-es v1.1(Carrino et al., 2020)and GermanQuAD(Möller et al., 2021a).

While MRC datasets exist for a variety of languages, a majority of these systems can only process popular languages such as English. Several MRC datasets also exist in other low-resource languages, such as Vietnamese ((Nguyen et al., 2020)) and Japanese ((So et al., 2022)), with many relying on Wikipedia articles.

Our focus here is on *Tigrinya*, a Ge'ez Script language with more than 10 million native speakers in Tigray, Ethiopia, and Eritrea (Abdelkadir et al., 2023). Tigrinya differs significantly from English regarding linguistic properties, including syntax, morphology, and typology (Gaim et al., 2023). While domain-specific labelled datasets are essential for evaluating a system's reading comprehension capacity, minimal effort has been put into creating annotated datasets in languages like Tigrinya leading to limited research on the same. Moreover, since Tigrinya is not the only language that suffers in MRC due to the scarcity of annotated datasets, we first investigate the state of MRC research in other such low-resource languages and discuss the major insights next.

Many existing low-resource language MRC datasets like UIT-ViQuAD (Vietnamese, Nguyen et al. (2020)), JaQuAD (Japanese, So et al. (2022)), and AmQA (Amharic, Abedissa et al. (2023)) rely on Wikipedia. On the other hand, some MRC datasets use Machine Translation and cross-lingual transfer techniques on existing English data for example the Persian ParSQuAD dataset (Abadani et al., 2021), and the Czech dataset (Macková, 2022). We argue that sourcing data from Wikipedia or translating from existing English data (especially in the case of low-resource languages) affects the dataset quality due to the following: *(i)* Wikipedia open source contributors in low-resource languages like Tigrinya are few, and their linguistic knowledge, relevance and authenticity can be easily questioned; *(ii)* Machine trans-lations from English to low-resource languages like Tigrinya have quality issues. Thus, further efforts are needed to create MRC datasets in low-resource languages like Tigrinya.

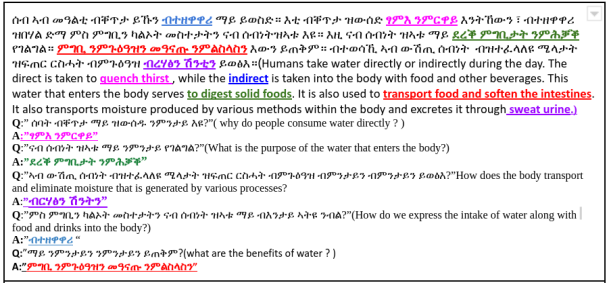

Figure 1: Sample of an educational paragraph (context) from the TɪɢQA dataset with 5 sample questions and labeled response spans.

Recently, complementary to our work, Gaim et al. (2023) proposed an MRC dataset and models for the Tigrinya language, a much-needed effort. However, they are yet to make their dataset publicly available. While they have focused on news article comprehension in Tigrinya, our work focuses on Tigrinya educational content comprehension and QA. To our knowledge, there are no previous works in domain-specific expert-annotated datasets in Tigrinya. Our domain-specific dataset can be a foundational building block for educational QA systems in Tigrinya, and also a valuable resource for MRC research in other low-resource languages. Figure 1 illustrates an example of our TɪɢQA dataset.

**Contributions.** We make the following major contributions:

*(1)* We provide a detailed empirical evaluation of machine translation (MT) models for translation, primarily when used for dataset creation. It underscores the necessity for more research into Tigrinya NLP and tailored QA systems for low-resource languages.

*(2)* We present TɪɢQA, an expert-annotated QA dataset in Tigrinya. To our knowledge, no other educational QA dataset exists in Tigrinya.

*(3)* We utilize TɪɢQA to evaluate human performance and analyze question and answer types with detailed language-related characteristics.

*(4)* Finally, our experiments with state-of-the-art multilingual pre-trained MRC models on TɪɢQA and comparison of their performance with human performance reveals various insights into span-based prediction in Tigrinya MRC.

TɪɢQA can be a good starting point for Tigrinya NLP models and can complement AI-based initiatives in the educational sector. The data is publicly available[1].

---
[1] https://github.com/hailaykidu/TigQA-Datasets

## 2. Related work

Numerous large annotated MRC datasets have been introduced to foster progress in question-answering and reading comprehension tasks. Most notably is SQuAD (Rajpurkar et al., 2016), which comprises 100K QA pairs that use Wikipedia articles as source contexts for crowd-sourced question generation and answer selection. Other datasets include Narrative QA (Kočiský et al., 2018), consisting of 4K QA pairs derived from 1572 stories gathered from books and movie scripts, MS-MARCO QA (Nguyen et al., 2016) providing around one million pairs of question-answers, and WikiQA (Yang et al., 2015), which offers 43k question-answer pairs. RACE (Lai et al., 2017) offers an educational domain QA dataset consists 100K items gathered from English exams, with four items per question. With human and AI contributions, TriviaQA (Joshi et al., 2017) presents 650K question-answer evidence triples gathered from Wikipedia, news websites, and articles. Despite the richness of these datasets, they primarily contain English text, mainly sourced from Wikipedia and other websites.

There are also multilingual (non-Tigrinya) QA datasets such as MLQA dataset (Lewis et al., 2020), which is a multi-way aligned extractive QA evaluation benchmark containing QA instances from over 12K question and answer samples in English and 5000 samples in six other languages such as Arabic, German, Spanish, Hindi, Vietnamese, and simplified Chinese while also encompassing various domains and languages. However, because of the generalization ability of multilingual and monolingual models, cross-lingual benchmarks like the Cross-lingual Question Dataset (XQuAD) (Artetxe et al., 2020), which comprises 240 paragraphs and 1190 question-answer pairs from SQuAD v1.1 translated into ten languages by professional translators, gains more attraction.

Monolingual MRC datasets beyond the English language are relatively rare. Albeit, some datasets designed to cover different aspects of QA are now available for various languages (Rogers et al., 2023), such as German (Möller et al., 2021a), Spanish (Carrino et al., 2019), Italian (Croce et al., 2018a), French (d'Hoffschmidt et al., 2020), Korean (Seungyoung et al., 2019), and Russian (Efimov et al., 2020). These datasets build upon the SQuAD blueprint with the most of them utilizing machine translation systems to translate from SQuAD to their target languages.

Furthermore, we investigate similar work in Tigriyna and the closely related Amharic language by Gaim et al. (2023) and Abedissa et al. (2023) respectively. The two languages share a similar alphabet based on the ancient Ge'ez script that dates back thousands of years and is considered part of the Semitic body of languages, including Arabic and Hebrew. However, they have distinct characteristics and are not interchangeable (Kaye, 2007). Note that Tigrinya has adopted many words from English (e.g., hotel, internet, or motor). It also borrowed some Italian words and shares many words with Amharic and Arabic (Tadross and Taxlu, 2017). Like other languages, Tigrinya varies in vocabulary and pronunciation by region. In the southern part of Tigray, the language does have interchangeable words between the two and differs somewhat from that found in Eritrea or northern parts of Tigray. To communicate effectively within this region, knowledge of Amharic is also helpful, which one can use all over Ethiopia (Tadross and Taxlu, 2017; Sahle, 1998). When it comes to reading and writing the language, the spelling of words tends to be more loosely defined than in English, partly due to regional differences in pronunciation. Also, some words and phrases vary between Tigray and Eritrea and even within the same region. This is similar to how Americans, English, and New Zealanders have a slightly different lexicon (Tadross and Taxlu, 2017).

While still relatively unexplored, there has been increasing research and interest directed towards Tigrinya in recent years. For instance, Yohannes and Amagasa (2022) developed a method of recognizing named entities for Tigrinya, while Tela et al. (2020) constructed a sentiment analysis dataset for Tigrinya language. Similar to our work, Gaim et al. (2023) proposed an MRC dataset and models for the Tigrinya language. However, the mentioned model and datasets are not yet accessible to the broader public, and the data source employed for this research is from news articles that are unsuitable for the educational domain. The same holds for Abedissa et al. (2023), who primarily relied on Wikipedia as a source for their Amharic QA dataset.

Although Wikipedia serves as an open-source knowledge repository for many languages and communities (Dzendzik et al., 2021), there is a disparity in the quality and quantity of content based on the languages. For example, Tigrinya suffers from poor quality as well as a small number of articles. Additionally, the content available in these languages is sometimes not contextually relevant and does not perfectly suit educational domain needs. Our proposed expert-annotated Tigrinya Question Answering dataset TɪɢQA is, therefore, a good starting point for different natural language tasks requiring question-answer sets. It can be used to train conversational AI systems (Zaib et al., 2022) tailored explicitly for the education sector to provide relevant and informative answers, thus improving the learning experience for students while also serving as a valuable resource for training machine learning models (Croce et al., 2018b). Such QA systems can also facilitate more efficient answer searching for students, allowing them to ask questions in their native Tigrinya language and receive accurate and targeted information. By condensing the content and presenting the most relevant information, students can quickly grasp the main points and comprehensively understand the topics covered in the TɪɢQA dataset.

## 3. TɪɢQA Dataset Collection

Even though multilingual data collections, such as Wikipedia, do exist for many languages (Yu et al., 2022), finding reliable data in Wikipedia for Tigrinya remains a significant problem due to the lack of trustworthy open-source contributors in Tigrinya. Additionally, the content in a language like Tigrinya is sometimes not contextually relevant and does not perfectly suit educational domain needs. Moreover, as discussed in Section 1, our analyses of machine translation for creating SQuAD-like datasets in Tigrinya indicate the quality of translation is an issue, especially when the target language is a low-resourced language like Tigrinya. Therefore, we focus on creating quality QA datasets annotated by experts and continuously evaluate them in the developmental process to enhance model performance. Unfortunately, unlike other languages, the Tigrinya language has no publicly available standardized annotated corpora akin to Treebank-3 (Marcus et al., 1999) or PropBank (Kingsbury and Palmer, 2002).

The collection of TɪɢQA was inspired by insights from existing datasets and our subject experts from the educational domains.

### 3.1. Creating Pages

To obtain high-quality and standardized data, we selected Tigrinya and Biology books from the Ethiopian Ministry of Education[2] used in elementary school (grades four and five) and high school (grades ten and eleven) to ensure both educational relevance and authenticity.

Initially, we collected ten books from all grades. Then, we selected five books based on their diverse reading topics. Four books are used for "Tigrinya," and one is for biology subject courses. These books vary in the number of pages and topics covered. We then divided them into two modules, represented by TɪɢQA-E and TɪɢQA-H. TɪɢQA-E consists of material for grades four and five, while TɪɢQA-H consists of books for grades ten and eleven (see Table 1). We only had scanned pages of the chosen books at our disposal, which are not machine-readable. These books included various elements, such as pictures, tables, and images. To convert the scanned books into editable, searchable digital text, we utilized an optical character recognition algorithm specifically usable for extracting Ge'ez Script (Tigrinya) texts called Tesseract[3] – an open-source OCR engine developed at HP between 1984 and 1994 (Smith, 2007).

---
[2]https://ethiopialearning.com/content/library

The scanned copy consisted of 800 draft pages before processing. However, we carefully cleaned the data to get well-structured paragraphs by removing unnecessary page numbers, figures, and tables. In addition, we discarded pages that contained duplicate sections. Finally, we extracted 455 pages and 537 paragraphs from 122 diverse topics, including climate, social sciences, culture, history, health, and business, which professional experts carefully predesigned. Furthermore, we created elementary school and high school modules based on the grade level of the book (see Table 1).

### 3.2. Selection of Annotators

The selection of the annotators was a meticulous process, including interviews to assess their expertise and teaching experiences. Four expert annotators were carefully selected based on their deep knowledge of the relevant subject matter for our research, ensuring they fully understood the content and vocabulary usage. In our interview, we investigate how experts recognize the usage of the exceptional character in Tigrinya. For example, there are characters we can write differently, but their reading is the same, for instance, θ and ጸ (both read as "Tse").

Additionally, annotators were required to reside in Ethiopia, where our research is focused, and should currently be teaching courses that fit the local curriculum and educational context. We believe that localized expertise is valuable for creating accurate annotations that are culturally and contextually relevant and suit the academic domain, making our annotation process creative and our data collection method reasonable and rigorous compared to datasets based on Wikipedia or news articles (Gaim et al., 2023; Rajpurkar et al., 2016; So et al., 2022; Nguyen et al., 2020). Furthermore, we sought individuals with extensive teaching experience in the respective subjects, enabling them to provide valuable insights into the nuances of the content.

Inclusivity was another critical consideration in our annotator selection process. We considered Tigrinya's linguistic diversity and prioritized choosing annotators representing different dialects and language variations within the country. This approach helps us capture a more comprehensive and inclusive perspective in our annotations, acknowledging the rich tapestry of languages and cultures that exist within Ethiopia.

Although we had to rely on recommendations to find fitting annotators, our rigorous selection process ensured high-quality annotations.

### 3.3. Question Answer Annotation

Experts were required to create questions and answers by reading TIGQA paragraphs. They were given the two modularized grade levels, elementary and high school, according to the students reasoning difficulty level and were encouraged to phrase questions in their own words. We provide the experts with detailed annotation instructions (see Appendix A), including a sample paragraph, examples of practical and impractical questions, and answers based on that paragraph for their reference. Moreover, we continuously monitored the annotations and gave feedback wherever necessary. Furthermore, we provided technical support in weekly meetings to support their tasks. Based on this, they created self-sufficient question-answer pairs that did not require additional information to answer.

For this task, the experts were instructed to create relevant annotations of 10-12 pages per hour. Each page can have 2-3 paragraphs. In terms of compensation, they were paid 5€ per hour for the time required to complete the assigned paragraphs. We did not enclose any wage benchmark since we could not find relevant work to estimate the time needed. We consider electric power and internet interruption situations so they can work on the task at their own flexible time. Therefore, the time that passed for some annotators may be longer due to occasional power outages or any other reason in their home country. However, we approximated that the annotator should finish at least ten pages to be paid and agreed to pay 5€ per hour.

Experts should draw upon their extensive knowledge of the subject to formulate five self-sufficient questions and corresponding answer spans by reading the paragraph. The experts had to create questions by reformulating text sections independently instead of building on word-for-word copying from the paragraph. We continuously monitored the annotation process through weekly discussions. Finally, our annotators created a total of 2685 question-answer pairs.

We randomly split the annotated data into training, development and test sets of 407, 65, and 65 paragraphs, respectively. The corresponding training, development and test sets of TIGQA-E

---
[3] https://github.com/tesseract-ocr/tesseract

and TıgQA-H contain 283, 40, and 40 resp. 254, 25, and 25 paragraphs.

### 3.4. Additional Answer Collection

To further ensure the quality of TıgQA, we assigned a secondary answer formulation task to all expert annotators to obtain two additional answers from the development and test set questions in TıgQA-E and TıgQA-H modules. To accomplish this, we provided experts with the question and the corresponding paragraphs from the page without any pre-existing answers. The experts carefully analyzed the paragraphs and selected the shortest span that directly addressed the question. If a question could not be answered within any range in the paragraph, the experts were instructed to indicate it as such without providing a specific answer. This approach aimed to avoid overlapping answers and evaluate the dataset's quality based on the percentage of correctly answered questions.

Expert annotators were advised to maintain a speed of 5 questions (equivalent to one paragraph) in 3 minutes for this task. They were compensated at a rate of 5 euros per hour for the total hours required to complete the entire set of questions for quality assessment.

Throughout the annotation process, we collected 2685 unique questions, and then 3315 answers, i.e., development and test set questions had at least three ground truth answers. From these answers, we found 37(<2%) entries of questions submitted without a specific span due to errors, incomplete, unanswerable by the context, or having a wrong question. Seventeen were corrected by changing the questions and modifying the paragraph. Twenty were question formulations that were not relevant to the given context.

Examples: **(Context)** ገለ ካብ ኻዕቲ ዝተረኸቡ መረዳእታታት መዕናዕቲ ከምዝሕብሩዎ ኣብ 5ይ ክፍለ ዘመን ኣብ ኣክሱም ካብ ፲፣000፡ ጁ፣000 ዝበዕሕ ቁዕሪ ህዝቢ ይነብር ከም ዝነበረ ይሕብሩ። [Some excavations indicate that 10,000 to 20,000 people lived in Aksum in the 5th century.] **(Question)** ኣብ ኣክሱም ብዝተገበረ ኻዕቲ እንታይ ዓይነት መዓድናት ተረኺቦም፧ [What minerals have been discovered in the excavations in Aksum?]. This is an irrelevant question in the given paragraph.

Moreover, questions in Tigrinya can be made by starting with interrogative pronouns (Tadross and Taxlu, 2017). TıgQA dataset consists of many types of questions(see Figure 2). We automatically associated each question with the primary question type to which it belonged. We first analyzed each question in TıgQA and discussed it with Tigrinya linguistic experts to perform this analysis. However, in Tigrinya, question words vary a lot, so our experts manually annotate the type carefully. The analyses and discussions allow us to accurately categorize each question into the most relevant types. We provide examples of the questions with each categorical question type in Figure 2. Note that the expected answer types are beyond proper noun entities.

| English | Tigriyna | Proportion | Example |
|---|---|---|---|
| What | እንታይ | 28.2% | ኣብ ሓደ ክባቢ ዘውቱር ዝኾነ ክባቢ ኣየር **እንታይ** እዩ፣ [what do we call the long lasting weather of a particular area ?] |
| Why | ንምንታይ፤ ስለምንታይ | 16.4% | ተስፋይ **ንምንታይ** ትምህርቲ ከቋርጽ ደልዩ፣[Why did Tesfay want to quit school?] |
| How | ከመይ ጌርዲ | 14.6% | ሕመም ዓሰ ካብ ሰብ ናብ ሰብ **ከመይ** ይመሓላለፍ፣ [How is tuberculosis transmitted from person to person?] |
| Where | ኣበይ ፣ናበይ፣ካበይ | 11.9% | ናይ ዓለምና ኦሉምፒክ 2020 **ኣበይ** ተሳሊጡ፣[Where was the 2020 World Olympics held?] |
| Which | ኣየናይ ፣ኣየነይቲ | 10.1% | ኣየናይ እንስሳ እዩ ነቲ ፓርክ ፍሉይ ድምቀት ዝህቦ፣ [Which animal gives the park a special brightness] |
| Who | መን | 9.4% | ናይቲ ትምህርቲ ቤት ርእሰ መምህር **መን** ይብሃሉ ፣[Who is the principal of the school ?] |
| When | መዓዝ፤መኣዝ | 6.7% | እቲ ፈተና **መዓስ** እዩ ዝጅምር፣[ When does the test start?] |
| other | ጥቀስ | 2.7% | መንግስቲ ኣብዚ ሰሙን ካብ ዝገበርዎ ስምምዕነት ዝተወሰኑ **ጥቀስ**፣ [Name some of the agreements the government made this week?] |

Figure 2: Question type distribution in TıgQA dataset: grouped by interrogative words. The highlighted color implies the interrogative words in Tigrinya.

### 3.5. Addressing Bias

While it is challenging to guarantee the complete elimination of bias in dataset creation, acknowledging potential issues allows us to make efforts to mitigate some of them. Our dataset is inherently education domain-specific and is derived from student books. Despite our efforts to choose paragraphs that cover a diverse range of topics, we select annotators based on their expertise in the subject matter, linguistic skills, and teaching experience. We prioritize selecting annotators who can represent various dialects and language variations of Tigrinya. The annotators' team comprises both two men and two women. Then, during the evaluation, we add two men and two women.

## 4. Dataset Analyses

TıgQA was built by having subject experts write questions for a given paragraph sourced from a student book and extract a specific answer from the paragraph. We classify our questions based on the difficulty level in two modules, which are Elementary School (TıgQA-E)and High School (TıgQA-H).

### 4.1. Question Types and Answer

As we mentioned in section 3, the TıgQA dataset was sourced from public student books intended for two distinct age groups in Ethiopia: Elementary School students (9-13 years old) and High School students (15-18 years old). To account for the significant difficulty gap and reasoning level requirement between these two subgroups, TıgQA-E corresponds to the elementary school books, while TıgQA-H represents the high school books.

|              | TıgQA-E |     |      | TıgQA-H |     |      | TıgQA |     |      |
|--------------|---------|-----|------|---------|-----|------|-------|-----|------|
|              | Train   | Dev | Test | Train   | Dev | Test | Train | Dev | Test |
| No. Pages    | 200     | 15  | 15   | 204     | 25  | 25   | 404   | 40  | 40   |
| No. Paragraphs | 203   | 40  | 40   | 204     | 25  | 25   | 407   | 65  | 65   |
| No. Topics   | 49      | 10  | 10   | 31      | 11  | 11   | 80    | 21  | 21   |
| No. Questions | 1215   | 100 | 100  | 1070    | 100 | 100  | 2285  | 200 | 200  |

Table 1: TıgQA dataset statistics

Experts fully manually prepare TıgQA, even though it is a costly and time-consuming task to prepare it, it brings more accuracy and integrity to the annotation. As shown in Table2, from the two modules, the number of paragraphs and topics is higher in TıgQA-E than in TıgQA-H. Moreover, we classify the questions into eight question-type groups as you can see from Figure-2, From these, the dominant question types in the dataset are inquiries starting with *what* (እንታይ) and *why* (ንምንታይ፣ ስለምንታይ), collectively comprising 59.2%. Subsequently, question types such as *where* (ኣበይ), *which* (ኣየናይ ፣ ኣየነይቲ), and *who* (መን) constitute 31.4%. In contrast, question types like *when* (መዓዝ፣ መኣዝ) and *other* (ጥቀስ) contribute a smaller proportion of 9.4%. Answering the predominant questions *what* and *why* requires a profound comprehension of the rhetorical structure and nuanced descriptions. Responses to such questions typically involve entire clauses, often independent sentences, rather than mere phrases embedded within a context closely aligned with the query.

Our dataset properly suits educational domains since we prepared it from public textbooks carefully designed for educational purposes. During the creation process, we continuously validate each of our corpus characteristics, such as context, questions, answer length, and size of vocabulary; as shown in Table 2, the paragraph, answer, question length, and vocabulary size are a higher portion in TıgQA-H than TıgQA-E this indicates the TıgQA-H require more reasoning-type questions based on difficulty at each grade level.

| Dataset | TıgQA-E | TıgQA-H | TıgQA |
|---------|---------|---------|-------|
| #Paragraph Len | 234 | 346 | 334 |
| #Question Len | 10.0 | 14.4 | 12.6 |
| #Answer Len | 3.1 | 5.3 | 5.0 |
| # Vocab Size | 14600 | 17601 | 32,201 |

Table 2: Statistics of TıgQA where Len denotes length and Vocab denotes Vocabulary TıgQA

Moreover, as we discussed in Section 3.2, we utilize local experts for annotations from individuals with linguistic expertise and subject teachers, which makes our dataset accurate, culturally, and contextually relevant for the educational domain; this helps to solve resource problems for low-resource languages of MRC/QA model training and machine translations in low resource languages. Our experts should aspire to formulate questioning using their language, avoiding direct word-for-word copying from the paragraph the annotator must draft on paper; they write it on the computer to prevent any mistakes. Finally, we provided technical support for using our annotation tool from Haystack [4], and following the provided support, the expert annotated the question and answer in the tool. This signifies that our dataset is exceptional and represents the first instance of subject matter experts' annotation in the low-resource language Tigrinya.

As shown in Listing 1, generated questions and their associated answer(s) are stored in the JSON format similar approach to (Rajpurkar et al., 2016)

Listing 1: Sample Generated Question and Answer from TıgQA dataset

```
{
"question":"ኣዝዩ ቆራር ኩነታት ኣየር ዘለዎ ኣየናይ
    ነባሪ ኣየር እዩ ?\n",
"id": 1157395,
"answers": [
    {
        "answer_id": 1051685,
        "document_id": 1715181,
        "question_id": 1157395,
        "text": "ዶጉዓ",
        "answer_start": 1409,
        "answer_end": 1412,
        "answer_category": "SHORT"
    }
],
"is_impossible": false
}
```

### 4.2. Reasoning Types of the Questions

The different levels of reasoning necessary to address TıgQA significantly impact the skills that models can acquire from the dataset. We strati-fied reasoning types using a variation on the taxonomy presented by (Trischler et al., 2016a; Lai et al., 2017; Mou et al., 2021).

---
[4]https://docs.haystack.deepset.ai/

**Word matching:** The question corresponds precisely to a section in the paragraph, which makes the answer obvious.

**Paraphrasing:** The question is implied or rephrased by a single sentence in the paragraph, and the answer can be extracted.

**Single-sentence reasoning:** The answer could be deduced from a single sentence in the paragraph, either by recognizing incomplete information or through conceptual overlap.

**Multi-sentence reasoning:** The answer necessitates inference by combining information scattered across multiple sentences.

For reasoning-type questions in TIGQA, one hundred question examples (drawn randomly from the respective development sets) were listed according to these types, and the results are compiled in Figure 3. For each, we show an example question with context that contains the answer span. Phrases relevant to the reasoning type are in bolded and colored. Some examples fall into more than one category, in which case, we defaulted to the more challenging type. As shown in Figure 3, word matching is the most accessible type and is the most significant subset of our datasets (27.2%). Paraphrasing constitutes (26.6%), and single-sentence and multi-sentence reasoning comprise 24.3% and 21.9%, respectively.

| Paraphrasing | Examples | Freq. |
|---|---|---|
| Word Matching | Q: ኣታ ድሙ ኣበይ እያ ነታ ኣንጭዋ ሃዲናታ፧ [Where did the cat chase the mouse?]<br>C: ኣታ ድሙ ነቲ ኣንጭዋ ብገር እቲ ጀርዲን ኣሳጊጋታ። [The cat chased the mouse across the garden.] | 27.2% |
| Paraphrasing | Q: ማይ ኣብ ምንታይ ሙቐት እዩ ዝፈልሕ፧ [What temperature does water boil at?]<br>C: ማይ ኣብ 100 ዲግሪ ሴንቲግሬድ ይፈልሕ። [Water boils at 100 degrees Celsius.] | 26.6% |
| Single-Sentence Reasoning | Q: ዮሃንስ ክንደይ ኣፕል ተሪፍዎ ኣሎ፧ [How many apples does John have left?]<br>C: ዮሃንስ ሓሙሽተ ኣፕል ኣለዎ። ንሳራ ክልተ ይህቦ። [John has five apples. He gives two to Sarah.] | 24.3% |
| Multi-Sentence Reasoning | Q: ቶሚ እንታይ ዓይነት ስፖርት ኣየ ዘዘውትርን፣ ኣበይከ እዩ ዝለማመዶ፧ [What sport does Tommy enjoy, and where does he practice it?]<br>C: ቶሚ ኩዕሶ እግሪ ይፈቱ እዩ። ኣብ ስፖርታዊ ኣዳራሽ ንሱዓታት ኣብ ልምምድን ኣካላዊ ምንቅስቓስን የሕልፍ። [What sport does Tommy enjoy, and where does he practice it?] | 21.9% |

Figure 3: In 100 randomly chosen samples from the TIGQAdataset development set, examples of questions alongside contexts containing the answer span relationships. Words pertinent to the reasoning and the selected answer type are highlighted. "Q" denotes the question, while "C" signifies the context.

## 5. Empirical Evaluation

### 5.1. Human Performance

We engaged four new experts to answer questions to measure human performance in development and test sets in TIGQA. As discussed in Section 3.4, each question in the development and test sets has at least three answers; while (Rajpurkar et al., 2016) adopts the second answer as the prediction, (Gaim et al., 2023) employs the third answer as a prediction. Moreover, (d'Hoffschmidt et al., 2020) and (Nguyen et al., 2020) compute the average by considering each of the answers as a prediction. We specify the first answer to each question as the human prediction while retaining the remaining answers as ground truth annotations. We used two evaluation metrics, exact match (EM) and F1-score, to evaluate the performances of MRC models on our dataset. We found that human performance scores on the development and test sets are 87.6 % EM and 92.4% F1 in TIGQA. Notably, disparities primarily arise from spare tokens in the answer spans rather than substantial discrepancies regarding the answers.

### 5.2. MT Error analyses

Although the translation of the existing English dataset can advance artificial intelligence (AI) capabilities for underrepresented languages like Tigrignya, it is crucial to realize the challenges with this task. We found three publicly available MT systems from English to Tigrigna and vice versa.

We first examine the feasibility of using machine translation (MT) to create a Tigrinya dataset similar to SQuAD. Therefore, we investigated different approaches by translating a line and a comprehension in both directions and then comparing auto and manual translations. Finally, we categorize the errors. Various translation methods are used to compare the quality of translations with human translations. For this task we used a sample of 150 triplets extracted from SQuAD (Rajpurkar et al., 2016) and an additional 50 Tigrinya triplets manually created following the SQuAD (Rajpurkar et al., 2016) format from the student textbook. The aim was to translate these pairs using three publicly available MT systems[5,6,7]. We found that public Google Translate had fewer errors than the two then as shown in (Appendix Tables 4 and 5) for sample translation of auto and manual, we carefully analyzed by language experts, and auto-translation(MT) faces difficulties translating proper nouns, vocabulary, syntax, errors of Omission, and untranslated words or phrases, which leads to a loss of context and meaning of the paragraph, which is difficult for the reader to understand. We categorize the error types into three main classes (Untranslated, Omission, and Mistranslation) based on the MQM-DQF typology, assigning specific terminology to each error type. We refer readers (Lommel, 2018) and (Abdelkadir et al., 2023) for a detailed explanation of the error types.

---
[5]https://lesan.ai/
[6]https://translate.google.com/
[7]https://www.tigrinyatranslate.com/main/

**Mistranslation:** This is the most common translation issue in current systems. These are commonly terminologies that could be technical, e.g., oxide. We observed many such errors in which a system translates a given terminology by taking a part of the source token. Another common type of mistranslation is words having different meanings depending on context (Appendix Tables 4 and 5 for detailed examples). Finally, we observed many occurrences of words translated with their antonyms (kick-off to start is translated as to finish).

**Omission:** The second most prevalent type of error is Omission. The primary kind of Omission is cases where current systems leave out an expression at the start, middle, or end of a sentence. Usually, in distribution by domain, the systems translate the main idea of a sentence and leave out supportive clauses. We observed in some systems, complete sentences are omitted from a snippet.

**Untranslated:** The other common type of error is untranslated tokens. These are usually abbreviations that refer to measurements (GWH) and named entities, e.g., political party names, currency, etc.

Finally, we evaluated the pre-trained model followed by (Devlin et al., 2019), a state-of-the-art QA system, using auto-translated and manually translated questions and paragraphs as input. The results revealed variations in the model's predictions for both inputs. These variations underscore the influence of translation quality on the model's performance in question-answering tasks. Translation accuracy directly affects the model's ability to predict correct answers (see Appendix table 7). This motivates us to prepare an expert-annotated benchmark dataset.

### 5.3. Baseline Models

**AfriBERTa$_{Base}$:** proposed by (Ogueji et al., 2021a) is a pre-trained multilingual language model with around 111 million parameters. The model has been shown to obtain competitive downstream performances on text classification and Named Entity Recognition in several African languages, including Tigrinya. We evaluate the model as a base for the first baseline.

**DrQA:** Chen et al. (2017) developed a simple but effective neural network-based model for the MRC task. The DrQA Reader achieved good performance on multiple MRC datasets (Rajpurkar et al., 2016; Reddy et al., 2019). Thus, we re-implement this method into our dataset.

**XLM-R:** which was proposed by (Conneau et al., 2020) a strong methodology pretraining multilingual language models at scale leads to significant performance gains for a wide range of cross-lingual transfer tasks. The model outperforms multilingual BERT (mBERT) on various cross-lingual benchmarks, including XNLI, MLQA, and NER. In this paper, we evaluate XLM-RBase and XLM-RLarge on our dataset.

## 6. Experimental Analysis

Our experiment specifically utilized the DrQA and AfriBERTa$_{Base}$ models, renowned for their effectiveness in natural language processing tasks. DrQA is established for Question answering, while AfriBERTa$_{Base}$ is tailored to address linguistic nuances in African languages like Tigrinya. Both models were finetuned using the HuggingFace transformers library[8] and the NVIDIA T4 GPU on Google Collaboratory. Following (Nguyen et al., 2020) method, we integrated various pre-trained word embeddings, including Word2vec, fastText, ELMO, and BERTBase. We use AdamW optimizer with default settings, a learning rate 3e–5, and a batch size of 16 were employed. Training extended over eight epochs, with a maximum sequence length of 334 tokens. While questions remained unaltered during preprocessing (maxing at 128 tokens), contexts can truncated to meet the model's sequence length requirements. This comprehensive setup facilitated a detailed evaluation of DrQA and AfriBERTa$_{Base}$ models' performance in Tigrinya question-answering tasks.

Listing 2: Prediction Sample
```
prediction = pipe.run(
  query = "አፍም ብዝኽፈትሉ ጀንጀ ምንባብን
           ምዕሓፍን ንምብቃዕ ዝተዳለወ ስርዓተ
           ትምህርቲ እንታይ እዩ?"
  params = {
    "Retriever": {"top_k": 5},
    "Reader": {"top_k": 5}
  }
)
```

### 6.1. Evaluation results

Improvement in low-resource languages has been made by introducing MRC/QA datasets. As part of our evaluation shown in Appendix Table 6, our dataset is unique due to its expert annotations, making it particularly suitable for educational contexts and culturally relevant to local use cases. In contrast, other datasets rely on sources such as Wikipedia and news articles, utilizing crowd workers instead.

---
[8] https://huggingface.co/

|  | EM | | F1 | |
|---|---|---|---|---|
| **Model** | Dev | Test | Dev | Test |
| DrQA + BERT | 46.71 | **52.10** | 56.08 | **60.03** |
| DrQA+ ELMO | 34.52 | 38.45 | 36.06 | 40.01 |
| DrQA+ fasrText | 37.73 | **42.38** | 51.03 | **58.08** |
| DrQA + Word2vec | 25.71 | 30.82 | 48.00 | 52.08 |
| AfriBERTa$_{Base}$ | 36.04 | **47.43** | 52.08 | **60.02** |
| XLM-R$_{Large}$ | 59.04 | **66.56** | 70.2 | **84.34** |
| XLM-R$_{Base}$ | 46.26 | 46.28 | 56.81 | 68.12 |
| Human Performance | 87.16 | 92.24 | 86.2 | 94.43 |

Table 3: Human and model performances on the Dev and Test sets of TIGQA

As shown in Table 3, the comparative performance of our models is compared against human performance in both our dataset's development and test sets. Regarding Exact Match (EM) and F1-core scores, XLM-R$_{Large}$ demonstrates significantly superior performance to the other models. Despite this, it needs to catch up to human performance levels. Specifically, the model achieves an F1 score of 84.34% on the test set. However, its Exact Match accuracy is 66.56%, considerably lower than its F1 score. This suggests that while the model identifies relevant answers, it struggles to precisely match human responses, indicating room for improvement in aligning its outputs more closely with human performance benchmarks.

### 6.2. Effect of Coverage Analyse

During our evaluation of MRC model on our dataset, we observed performance discrepancies linked to the length and input type. Specifically, we scrutinized the impact of question length, answer length, passage length, Question type, answer type, and reasoning type on performance. For instance, we noticed a decline in performance as answer length increased excessively. Conversely, the model demonstrated satisfactory performance when dealing with straightforward entity-type answers. However, there were discernible fluctuations in performance, particularly with higher-level reasoning-type questions. For instance, questions involving "why" faced particular difficulty. In summary, our analysis revealed that the MRC model's performance is affected by questions, answers, passage lengths, and the types of questions, answers, and reasoning involved. These findings emphasize the importance of considering various aspects of input data when assessing and enhancing the performance of MRC models on Tigrinya datasets.

## 7. Conclusion and Further Work

This paper introduced a new expert-annotated dataset (TIGQA) tailored for reading comprehension and question-answer tasks in the Tigrinya language within the educational domain. The dataset was curated by extracting Ge'ez Script (Tigrinya) documents from scanned student textbooks using the Tezeract OCR engine. Our work not only advances the imperative task of resource creation for low-resource African languages like Tigrinya but also forges a path toward innovative educational applications and a plethora of research avenues. The dataset's quality and thematic breadth position it as a valuable asset, both for advancing the field of natural language processing in low-resource language contexts and fostering enhanced educational interactions between students and their mentors. In our upcoming research, we plan to expand the dataset's size and conduct additional experiments by fine-tuning various models on the different tasks provided in the dataset. Moreover, the dataset can be used for any NLP model pieces of training since experts carefully prepare it, but we also observed that preprocessing for Tigrinya, like tokenization, affects model performances. By making TIGQA publicly available, we not only provide a new benchmark for reading comprehension tasks but also support the reuse and further expansions of our dataset.

## Acknowledgements

We want to thank the reviewers' comments, which have helped improve the quality of our work. In addition, we would like to thank our expert annotators for their cooperation.

# Appendix A. Annotation Guidelines and Data Overviews

### Annotation Guidelines

**Description of the document:** Before you do anything, Read the structure and meaning of the document as the first instruction. The corpus is Tigriyna. Data was collected from March to June 2023: The meaning of the columns below.

**Id:** is the assigned identifier to each row instance

**Category:** This is the category of the context with which topic the context belongs

**Sample Q1:** The sample corrects and incorrects questions from the context.
**Sample A1:** Correct and inaccurate sample of answer to the question.
**Source:** The Source of the document

Question Answering (QA) is vital in education, representing the primary interaction between instructors and students. This paper introduces the first taxonomy and annotated educational corpus of questions to help analyze student responses. The dataset can be in approaches that classify questions based on the expected answer types or can also extractive answers from the documents retrieved based on the query. This dataset can used to train other NLP monolingual and multilingual models.

Guidelines for creating QA: These are for research purposes. Follow the given template and guidelines; please ask me for clarification if you have any questions or concerns.

**1.** Read and understand the given context: Before creating question and answer pairs, carefully read and understand the context or passage provided. Take note of critical concepts, ideas, and details.
**2.** Identify important information: try to identify the most critical information in the context to create the question answer in your own words.
**3.** Use clear language: Write questions and answers using clear and concise language that is easy to understand. Avoid using overly technical or complex language that may be difficult for readers to understand.
**4.** Be specific: Write questions focused on a particular aspect of the context. Each answer should be a clear span of text from the given context that directly answers the question.
**5.** Use extractive question types: Use questions that can be answered as text spans from the context. This can help to evaluate the reader's understanding of the context quickly.
**6.** Ensure accuracy: Ensure each question and answer span is accurate and factually correct. Verify the accuracy of the information in the context before creating questions and answers.
**7.** Ensure relevance: Ensure each question and answer pair is relevant to the context. Avoid including questions and answers irrelevant to the context or too narrow in scope.
**8.** Consider the intended audience: Consider the intended audience(grade-level students) when creating questions and answers. Write questions and answers that are appropriate for the reader's level of knowledge and understanding.
**9.** Review and revise: Review and revise questions and answers for accuracy, clarity, and relevance. Make necessary changes to ensure the questions and answers meet the standards according to the existing curriculum.
**10.** ensure that the question and answer pairs in the dataset accurately and effectively test the reader's understanding of the context, using span-based question types to evaluate their knowledge and reading comprehension.

| | |
|---|---|
| Original Reference Paragraph (SQuAD) | Beyonc Giselle Knowles-Carter (/ bee-YON-say) (born September 4, 1981) is an American singer, songwriter, record producer and actress. Born and raised in Houston, Texas, she performed in various singing and dancing competitions as a child and rose to fame in the late 1990s as lead singer of R&B girl-group Destiny's Child. Managed by her father, Mathew Knowles, the group became one of the world's best-selling girl groups of all time. Their hiatus saw the release of Beyonc's debut album, Dangerously in Love (2003), which established her as a solo artist world-wide, earned five Grammy Awards and featured the Billboard Hot 100 number-one singles "Crazy in Love" and "Baby Boy" |
| Autotranslation to Tigrinya | ቢዮንክ ጄዘል ኖውልስ-ካርተር (bee-YON-say) (4 መስከረም 1981 ተወሊዳ) ኣመሪ-ካዊት ደራፊት፡ ደራሲት ድርፊ፡ ኣፍራዪት መዝገብን ተዋሳኢትን እያ። ኣብ ሂዩስተን ቴክሳስ ተወሊዳ ዝዓበየት ንሳ፡ ንእስነታ ኣብ ዝተፈላለየ ውድድራት ደርፍን ሳዕስዒትን ተዋሲኣ፡ ኣብ መወዳእታ 1990 ታት ድማ ከም መሪሕ ደራፊት ናይ R&B ጓል-ጉጅለ ደስቲኒስ ቻይልድ ኮይና ናብ ዝና ደየበት። ብኣቦኣ ማቲው ኖውልስ እትመሓደር ዝነ-በረት ጉጅለ፡ ሓንቲ ካብተን ኣብ ኩሉ ግዜ ዝሽየጣ ጉጅለታት ኣዋልድ ዓለም ኮይና። ዕረፍቶም ድማ ናይ ቢዮንክ ናይ መጀመርታ ኣልቡም Dangerously in Love (2003) ክትዝርጋሕ ከላ፡ ኣብ መላእ ዓለም ከም ሶሎ ኣርቲስት መስሪታ፡ ሓሙሽተ ሽልማት ግራሚ ረኺባ፡ ኣብ ቢልቦርድ ሆት 100 ቁጽሪ ሓደ ንጽል ደርፋታት "Crazy in Love" ን "ን ዘርኣየት። ህጻን ወዲ" |
| Human Translation from Autotranslated Tigrinya to English | Beyonk Jizel nowles Karter (bee-YON-say) born September 4, 1981) is an American singer, song author, record manufacturer and actress. In Hiyusten texas born and grow up she, her childhood participated in the different computation of song and dancing, in the late 1990s as a leader English songs of R&B girl group Destiny's child climbed to fame. the group administered by her father Mathew Knowles, became one of the sold group girls in the world. their vacation also of biyonk first album Dangerously in Love (2003), while establishing, in the whole world as a solo artist. She earned five Grammy Awards, in billboard hot 100 number one single sings "Crazy in Love" and for which showed. Baby boy" |
| Error Analysis | When comparing the outputs of auto-translated and manually translated text, several challenges become evident. Proper nouns, such as names or specific terms, pose difficulties for the MT. It struggles with translating them accurately, resulting in errors or omissions. Additionally, vocabulary and syntax may not be rendered correctly, leading to a loss of context and meaning in the paragraph. Untranslated words or phrases further contribute to the reader's difficulty in understanding the text. In some cases, the MT fails to separate or indicate specific details like the year count, such as distinguishing "1981 G.c" from the Ethiopian calendar. Furthermore, there can be instances of misinterpretation and the unnecessary addition of punctuation, further hindering comprehension. |

Table 4: Example of Paragraph-wide Translation Errors

| | |
|---|---|
| Original Reference Paragraph (TɪɢQA) | ነባሪ ኣየር ፦ ነባሪ ኣየር ኣብ ሓደ ከባቢ ዝውቱር ዝኾነ ኩነታት ኣየር እዩ። እዚ ምስ ስነ ምድራዊ ኣቀማምጣ ከባቢ ቀጥታዊ ርክብ ኣለዎ። እዚ ኩነታት ሓደ ከባቢ ካብ ዝግለፀሎም መዳያት ሓደ እዩ። ነባሪ ኣየር ደጉዓ፣ ሓውሲ ደጉዓን ቆላን ተባሂሉ ኣብ ሰለስተ ይክፈል። ሕድሕድ ነባሪ ኣየር ናይ ባዕሉ ዝኾነ ብራኽን መጠን ዋዒን ኣለዎ። " ደጉዓ ዝኾኑ ቦታታት ካብ ፀሓ ባሕሪ ንላዕሊ ካብ 2,500-4,000 ሜትር ዝኸውን ብራኽ ኣለዎም። ደጉዓ ኣዝዩ ቆራርን ኣስሓይታ ዝበዝሐን ኩነታት ኣየር ኣለዎ። እዚ ነባሪ ኣየር ከም ሩዝ፣ ስርናይ፣ ዓይኒዓተርን ዓተርን ዝበሉ ዘእትቲ ከም ሰዓን ኣውሱዳን ዝበሉ ቅመማትን ንምፍራይ ምቹው እዩ። ብተመሳሳሊ ደጉዓ ከም ኣፍራስን ኣባጊዕን ንዝበሉ እንስሳት ዝሰማማዕ ነባሪ ኣየር እዩ። ሓውሲ ደጉዓ ድማ ባህሪይ ነባሪ ኣየር ደጉዓን ቆላን ኣለዎ። እዚ ነባሪ ኣየር ዘለዎም ከባቢታት ካብ ፀሓ ባሕሪ ንላዕሊ ካብ 1,500-2,500 ሜትር ብራኽ ኣለዎም። ሓውሲ ደጉዓ ማእኸላይ ሙቐት ኣለዎ። ኣብዚ ካባቢ ከም ዳጉሻን ጣፍን ዝበሉ ዘራእቲ ብዝበለፀ ይሰማምዖም። ካብ 500-1,500 ሜትር ብራኽ ዘለዎ ነባሪ ኣየር ቆላ ይበሃል። እዚ ነባሪ ኣየር ኣዝዩ ምዉቕ ዝኾነ ኩነታት ኣየር ኣለዎ። ቆላ ንከም ምሽላ፣ ምሽላ ባሕሪ፣ ስሊጥን ኒሁግን ዝበሉ ዘራእቲ ይሰማማዕ። ከም ኣጣልን ኣግማልን ዝበሉ እንስሳ ዘቤት ንምፍራይ ቆላ ይምረፅ። |
| Autotranslation to English | Permanent air: Residential air that is common in an area It's the weather. This geological Location has a direct bearing on the environment. This situation is one of the areas where they are described. Aspects are the same. Resident Air Degua, Mixed Degua and Plain in three It is paid. Each resident has his own air Any height and temperature. Degua areas are at an altitude of 2,500-4,000 m above sea level. Degua has a very hot and sunny climate. This permanent atmosphere is suitable for the production of crops such as rice, wheat, millet and barley and spices such as sesame and awsuda. Similarly, degua is a permanent climate suitable for animals such as horses and sheep.It is called a permanent air plain at an altitude of 500-1,500 m. This permanent air has a very temperate climate. Plain suits crops like millet, seaweed, celery and nihug. The plains are preferred for the production of domestic animals such as goats and camels. |
| Human Translation to English | Climate: Climate is the long-lasting weather of a particular area. It has a direct connection with the geographical characteristics of a region. This one way of describing a certain place. Climate is divided into three categories: Highland, semi highland and lowland. Every climatic region has their own elevation and temperature margin. Highland regions are those located from 2,500-4,000m above sea level. This climate has extremely cold and frosty weather conditions. This type of climate is ideal to grow crops like rice, wheat, pea and chickpeas and spices like basil and black seed. Similarly, it is convenient to farm animals such as horses and sheep. Semi highland has a mixture climate of lowland and highland. It is designated to places with elevation from 1500-2500m above sea level. Semi Highland has moderate temperature. It is ideal for grains like Teff and finger millet. Climate with elevation from 500-1500m above sea level is called lowland. Corn, sesame, millet and oil seeds can grow in this type of climate. Moreover, regions with this climate are preferable to raise domestic animals such as goats and camels. |
| Error Analysis | The meaning and context of the reading are lost when auto-translated from Tigrinya to English, failing to recognize different words or phrases. For example, the main topic of the reading word or phrase is about "ነባሪ ኣየር", which is translated as "Permanent air". However, the correct translation is "climate". Words like "ደጉዓ" are falsely translated as Degua, though the correct translation should be "Highland". "ቆላ" is mistranslated as plain, but the correct translation should be "lowland". Similarly, the word "nihug" is not correctly translated. The correct translation should be "oil seeds". Even the meaning of entire sentences is lost in the autotranslation. For instance, Google MT translates the sentence: "ደጉዓ ኣዝዩ ቆራርን ኣስሓይታ ዝበዝሐን ኩነታት ኣየር ኣለዎ።" as "Degua has a very **hot and sunny climate.**" This does not reflect the provided information at all. The correct translation would be, "The Highland climate has extremely cold and frosty weather conditions". Moreover, the MT system does not recognize local terms like "Teff", which further decreases the quality of translations. |

Table 5: Manual and Automated Translations of a TɪɢQA Sample Paragraph

| Dataset | Language | Span-based | Professionally Annotated | Sourced from Student Books | Suited for Educational Domain |
|---|---|---|---|---|---|
| TiQuAD (Gaim et al., 2023) | Tigrinya | X | - | - | - |
| AmQA (Abedissa et al., 2023) | Amharic | X | - | - | - |
| UIT-ViQuAD (Nguyen et al., 2020) | Vietnamese | X | - | - | - |
| JaQuAD (So et al., 2022) | Japanese | X | - | - | - |
| ParSQuAD (Abadani et al., 2021) | Persian | X | - | - | - |
| Czech SQuAD (Macková, 2022) | Czech | X | - | - | - |
| IDK-MRC (Putri and Oh, 2022) | Indonesian | X | - | - | - |
| **TɪɢQA (Our dataset)** | Tigrinya | X | X | X | X |

Table 6: Comparison of TɪɢQA with existing low-resource MR/QA datasets. Our dataset is unique because it is fully annotated by experts, which is suited for educational domains and contextually and culturally relevant to the local use cases; others use Wikipedia and news articles as sources and employ crowd workers.

Table 7: Sample question of the models input.

| Reference Tigrinya Question from TigQA | Auto Translation (GoogleMT) | Human Translation | Answer Span | Model Prediction (Right/Wrong) |
|---|---|---|---|---|
| ኣብ ሓደ ከባቢ፣ ነውሓር ዝጸንሕ ክንታት ኣየር እንታይ ኢዩ? | What is the typical weather in an area? | What do we call the long lasting weather of a particular area? | Climate | r-r |
| ይማማ በታ ከመይ ዓይነት ክንታት ኣየር ኣለዎ? | Do you have air in a dog-like place? | What kind of weather is it in highland areas? | extremely cold and frosty | w-r |
| ካብ ዓቐብ ባሕሪ ንላዕሊ ካብ 1500 ክሳብ 2500 ሜትር ብራኽ ዘለዎ እንታይ ኢልና ንጽውዖ ? | What do we call an altitude of 1500 to 2500 meters above sea level? | What do we call a place with elevation from 1500 -2500m above sea level? | semi highland | r-r |
| ቆላ ንኣየኖት እንስሳት ኣዩ ዝምችእ, ? | Which animals are suitable for the plain? | For which animal is the lowland convenient? | horses and sheep | r-r |
| ከም ዓይነተ-ርጉድ ዓተር፣ ዝበለ ዝራእትን ከም ስሰግን ኣውሊድን ዝበለ ቅመማቅመም ገምራሪ ምቻዉ፣ ነቢረ ኣየር ኣየናይ ኣዩ ? | Which is the suitable permanent climate for the production of crops like rice, wheat, millet, barley and cereals like sesame and oats ? | Which Climate is ideal to grow crops like rice, wheat, pea and chickpeas and spices like basil and black seed? | Highland | r-r |
| ነቢረ ኣየር ምስ ምንታይ፣ ኣዩ ርክብ ዘለዎ? | What does permanent air have to do with it? | With what relation does a climate have? | geographical characteristics of a region | w-r |
| ሐዳ ሐዳ ነቢረ ኣየር ናይ ባዕሉ ብኽ ኣንታይ ዓይነት ባህሪ ኣለዎ ? | What is the characteristic of each dwelling air that is its own? | What characteristics does every climatic region have? | elevation and temperature margin | w-r |
| ኣብ ቆላ ዝስማዕማዑም እንስሳ ዘቤት ኣነመን ኣዮም ? | Who are the pets I agree with in the valley? | Which domestic animals are preferred for lowlands? | goats and camels | r-r |
| ከም ዳጉሻን ጣፍን ዝበለ ዝራእታታት ኣበይ ይስማዕማዑም ? | Where do crops like dagusha and rice fit ? | Where is the ideal place for grains like Teff and finger millet? | Semi Highland | w-r |
| ቆላ ዝስማዕማዑም ዓይነታት ዝራእቲ ኣነመን ኣዮም ? | What are the crops that are suitable for the plain? | What seeds can grow in lowland climates? | Corn, sesame, millet and oil seeds | w-r |
| ካብ ዓቐብ ባሕሪ ንላዕሊ ካብ 500 ክሳብ 1500 ሜትር ብራኽ ዘለዎ ክንታት ኣየር ታይ ይበሃል ? | What is the name of the permanent air at an altitude of 500 to 1500 meters above sea level? | What is the climate at an elevation from 500-1500m above sea level? | lowland | r-r |
| ኣዚ ቆራር ክንታት ኣየር ዘለዎ ኣየናይ ነቢረ ኣየር ኣዩ? | Which resident has the hottest weather? | Which climate has extremely cold and frosty weather condition? | highland | r-r |